\documentclass{article} 
\usepackage{iclr2023_conference_tinypaper,times}

\usepackage{amsmath,amsfonts,bm}

\def\eqref#1{equation~\ref{#1}}

\def\1{\bm{1}}

\DeclareMathAlphabet{\mathsfit}{\encodingdefault}{\sfdefault}{m}{sl}
\SetMathAlphabet{\mathsfit}{bold}{\encodingdefault}{\sfdefault}{bx}{n}

\usepackage{hyperref}
\usepackage{url}
\usepackage{graphicx}
\usepackage{subcaption}

\title{User Modeling Challenges in Interactive AI Assistant Systems}

\author{Megan Su, Yuwei Bao\\
Department of Computer Science and Engineering\\
University of Michigan\\
Ann Arbor, MI 48109, USA \\
\texttt{\{meganysu, yuweibao\}@umich.edu} \\
}

\iclrfinalcopy 
\begin{document}

\maketitle

\begin{abstract}
Interactive Artificial Intelligent(AI) assistant systems are designed to offer timely guidance to help human users to complete a variety tasks. One of the remaining challenges is to understand user's mental states during the task for more personalized guidance. In this work, we analyze users' mental states during task executions and investigate the capabilities and challenges for large language models to interpret user profiles for more personalized user guidance.  
\vspace{-10pt}
\end{abstract}

\section{Introduction}
\vspace{-10pt}

In the digital age, there is immense potential for artificial intelligent (AI) assistant to guides users through complex tasks, from changing laptop batteries to piping frosting on a cake. One of the main challenges, however, lies in creating an interactive system that can not only understand which step the user is at, but can also detect user's mental states, such as frustration, familiarity with the task, detail-orientation, etc. Only when the model understands user's personalized preferences can it provide a user-friendly experience and promoting better task completion outcome.

In this work, we address this challenge by 1) extending the WTaG (\cite{wtag}) dataset to incorporate user mental profiles during the task, such as frustration, eagerness to speak, and detail-orientation, and 2) by investigating if existing large language models (LLM) can accurately interpret underlying users states. This work aims to enhance user experience with AI assistance for the next steps by tailoring responses to best accommodate user's emotional and mental dispositions.

\section{Dataset - User Modeling Analysis}
\vspace{-10pt}

\begin{figure}[b]
  \centering
  \begin{minipage}[b]{0.4\textwidth}
    \includegraphics[width=\textwidth]{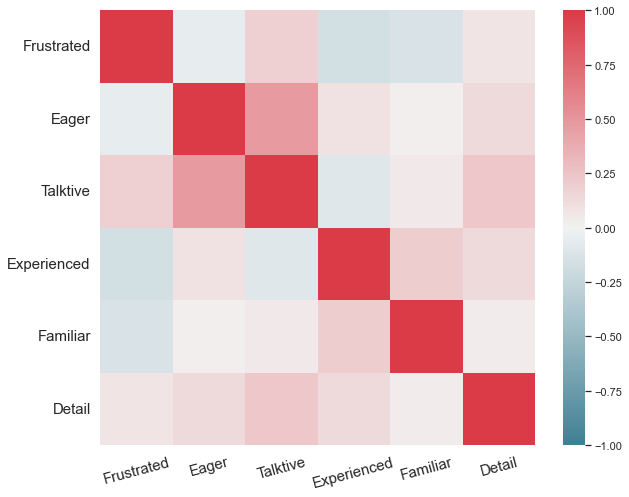}
    \caption{Feature Correlations}
    \label{corr}
  \end{minipage}
  \hfill
  \begin{minipage}[b]{0.59\textwidth}
    \includegraphics[width=\textwidth]{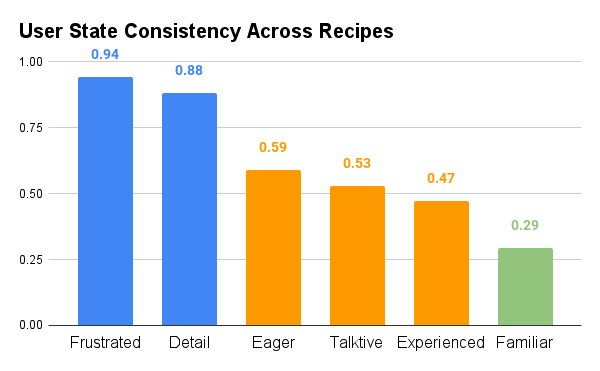}
    \caption{User Mental State Consistency Across Recipes}
    \label{consis}
  \end{minipage}
  \vspace{-10pt}
\end{figure}

AI Assistant systems such as HoloAssist (\cite{holoassist}), Ego How-TO (\cite{egohowto}), Google Assistant (\cite{googleassist}), Siri (\cite{siri}), and Alexa (\cite{alexa}) are designed to guide human users on various tasks. Some of which leverage augmented reality headsets to enable visual and verbal communications. In this work, we expanded the human instructor-user dialog from the WTaG (\cite{wtag}) dataset to include 6 categories of user profiles for each recording: \textbf{Frustration}, \textbf{Eagerness to ask questions}, \textbf{Talkative}, \textbf{Experience}, \textbf{Familiarity with tools}, and \textbf{Detail-orientation}. For each of the 55 recordings (across 17 users and 3 recipes), we annotated ``yes'' or ``no'' to each category above. A total of 330 data points were collected.

While choosing the 6 user profile categories, we try to cover a wide range of aspects to model users' mental states. We conducted a pair-wise Pearson correlation analysis to see how different features are related. Figure \ref{corr} reveals that other than ``Eager to ask questions'' and ``Talkative'', which are moderately correlated, most of the features depict a unique perspective of the user states. Other interesting observations include that users' frustration has a mild positive correlation with talkativeness and negative correlation with their experience, which suggests AI assistants could adjust their guidance according to user's proactive engagement or task-related experience level.

We also examined the user's mental state consistency across different recipe executions. For each user, the consistency is 1 if the user exhibited the same ``Yes'' or ``No'' to each profile category for all three recipes, and 0 otherwise. Among the results in Figure \ref{consis}, a high consistency was observed on users for both ``Frustration'' and ``Detail-orientation'', suggesting that AI assistants could transfer their user-specific accommodations across different tasks according to user's traits on attention to details, or external factors, such as user's mood for that particular day. ``Eager to ask questions'', ``Talkativeness'' and ``Experienced'' are moderately consistent for each user across recipes, suggesting a combination of both user-specific accommodation and recipe-dependent adjustment. ``Familiarity with tools'' have the lowest consistency, which indicates the AI assistant would need to understand user's proficiency on each tool used in different tasks for more situated guidance.

\section{User Modeling with Large Language Models}
\vspace{-10pt}

Several previous works (\cite{um3, um2, um1}) proposed using deep learning methods for user modeling. In this work, we investigate how large language models can interpret user's mental states during the tasks. Given the dialog history between the user and the instructor for each recording, we prompt ChatGPT\footnote{\url{https://openai.com/blog/chatgpt}; Version GPT3.5-turbo-0301.} to predict ``yes'' or ``no'' for each user profiling category. By comparing the F1 score, precision and recall of model's predictions with our annotations,  we gain insights into the model's ability to interpret users' needs and mental states. We set the temperature to zero, repeated the experiments 3 times and observed negligible variance.

\begin{figure}
        \begin{subfigure}[b]{0.33\textwidth}
                \includegraphics[width=\linewidth]{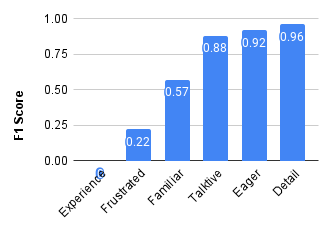}
                \caption{F1 Score}
        \end{subfigure}%
        \begin{subfigure}[b]{0.33\textwidth}
                \includegraphics[width=\linewidth]{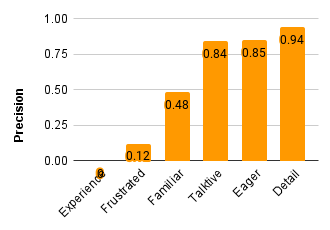}
                \caption{Precision}
        \end{subfigure}%
        \begin{subfigure}[b]{0.33\textwidth}
                \includegraphics[width=\linewidth]{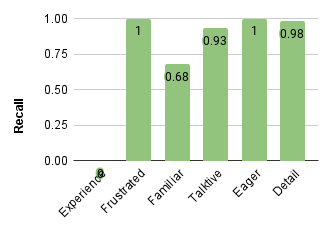}
                \caption{Recall}
        \end{subfigure}%
        \caption{ChatGPT User Modeling Prediction Performance}\label{results}
        \vspace{-20pt}
\end{figure}

Our results in Figure \ref{results} indicate LLM's high ability to detect when users are ``Detail-oriented'', ``Eager to ask questions'', or ``Talkative'', with F1 scores of 0.96, 0.92, and 0.88, respectively. However, the ``Frustrated'' category reveals a stark contrast, in which it frequently misclassifies non-frustrated behavior, yielding a high false positive rate in this category. With a precision and recall of zero, the ``Experienced'' attribute needs significant improvement, pointing to necessary steps to improve models' ability to understand user's knowledge and experience on the tasks.

\section{Conclusion}
\vspace{-10pt}

In this work, we analyze human users' mental states during task executions, and study how well can LLM based interactive AI assistant systems understand and accommodate for these personalized needs. We extended the WTaG dataset to incorporate 6 categories of user's mental states based on the dialog history, and challenge the LLMs to predict user states. Our analysis showed that users exhibited different levels of consistency across user profile categories and that significant improvements are needed for LLMs to understand several categories. 
Future work could include larger scale dataset analysis (e.g. \cite{holoassist, egoexo4d}), and investigate better prompting strategies, online user state detection and in-context adjustments, categorized fine-tuning, and multimodal signals for better user modeling.

\bibliography{iclr2023_conference_tinypaper}
\bibliographystyle{iclr2023_conference_tinypaper}


\end{document}